\documentclass[journal]{IEEEtran}
\IEEEoverridecommandlockouts
\usepackage{cite}
\usepackage{amsmath,amssymb,amsfonts}
\usepackage{algorithmic}
\usepackage{graphicx}
\usepackage{textcomp}
\usepackage{caption}
\usepackage{multirow}

\usepackage[ruled,linesnumbered]{algorithm2e}
\usepackage{xcolor}
\def\BibTeX{{\rm B\kern-.05em{\sc i\kern-.025em b}\kern-.08em
    T\kern-.1667em\lower.7ex\hbox{E}\kern-.125emX}}
\begin{document}

\title{CA-BERT: Leveraging Context Awareness for Enhanced Multi-Turn Chat Interaction\\
}



\author{
    \IEEEauthorblockN{Minghao Liu\IEEEauthorrefmark{1}, Mingxiu Sui\IEEEauthorrefmark{2}, Yi Nian\IEEEauthorrefmark{3}, Cangqing Wang\IEEEauthorrefmark{4}, Zhijie Zhou\IEEEauthorrefmark{5}}\\
    \IEEEauthorblockA{\IEEEauthorrefmark{1}Arizona State University, Tempe, AZ 85281, US.
    \\\{mliu107\}@asu.edu}
        \IEEEauthorblockA{\IEEEauthorrefmark{2}University of Iowa, Iowa City, IA 52242, US.
    \\\{suimingx\}@gmail.com}
    \IEEEauthorblockA{\IEEEauthorrefmark{3}Independently contributed to this work.
    \\\{ynian.4\}@gmail.com}
       \IEEEauthorblockA{\IEEEauthorrefmark{4}Ocean AI
    \\\{cwang\}@ocean-ai.io}
       \IEEEauthorblockA{\IEEEauthorrefmark{5}Independently contributed to this work.
    \\\{zhijiesabrinazhou\}@gmail.com}
}
\maketitle

\begin{abstract}
Effective communication in automated chat systems hinges on the ability to understand and respond to context. Traditional models often struggle with determining when additional context is necessary for generating appropriate responses. This paper introduces Context-Aware BERT (CA-BERT), a transformer-based model specifically fine-tuned to address this challenge. CA-BERT innovatively applies deep learning techniques to discern context necessity in multi-turn chat interactions, enhancing both the relevance and accuracy of responses.

We describe the development of CA-BERT, which adapts the robust architecture of BERT with a novel training regimen focused on a specialized dataset of chat dialogues. The model is evaluated on its ability to classify context necessity, demonstrating superior performance over baseline BERT models in terms of accuracy and efficiency. Furthermore, CA-BERT's implementation showcases significant reductions in training time and resource usage, making it feasible for real-time applications.

The results indicate that CA-BERT can effectively enhance the functionality of chatbots by providing a nuanced understanding of context, thereby improving user experience and interaction quality in automated systems. This study not only advances the field of NLP in chat applications but also provides a framework for future research into context-sensitive AI developments.
\end{abstract}

\begin{IEEEkeywords}
Meta-reinforcement learning, theoretical analysis, generalization bound, convergence guarantee
\end{IEEEkeywords}

\section{Introduction}
In the realm of natural language processing (NLP), the advent of transformer-based models such as BERT (Bidirectional Encoder Representations from Transformers) has significantly advanced the capabilities of text classification systems. These models have shown exceptional performance across a variety of NLP tasks by leveraging deep contextual representations. However, their application is often hampered by the need for extensive computational resources and large annotated datasets for fine-tuning. This research addresses these challenges by introducing the CCC-BERT model, a specialized variant of BERT fine-tuned for the task of context necessity classification in multi-turn chat environments.

Our study focuses on the practical application and fine-tuning of CA-BERT, a model initially pre-trained on a diverse corpus and subsequently adapted to identify whether a given text input in a chat requires additional context to be understood. This capability is critical for enhancing the efficiency of automated chat systems, where understanding the necessity of context can streamline interactions and improve response accuracy.

This paper outlines the architecture of CA-BERT, discusses its training on a novel dataset specifically curated for context classification, and evaluates its performance against standard BERT implementations. The contributions of this work are twofold: firstly, the adaptation of BERT to a niche but crucial aspect of chat-based systems, and secondly, the presentation of a methodology for efficiently training language models on specialized tasks without the need for expansive resource commitments.

\section{Background and Related Work}
The task of enhancing chatbot interactions through improved context awareness has seen considerable interest within the field of natural language processing. This section reviews relevant literature, focusing on advancements in transformer-based models and their applications in chat systems, with a particular emphasis on context sensitivity.
\subsection{Transformer Models in NLPs}Since the introduction of BERT announced in 2018, transformer-based models have revolutionized the landscape of NLP. Their ability to capture deep contextual relationships within text has made them the foundation for many subsequent innovations. Models such as RoBERTa and GPThave extended these concepts, focusing on varying aspects of model architecture and training approaches to enhance performance across diverse NLP tasks\cite{li2024dynamic}.
Further developments have included techniques such as attention mechanisms that allow models to focus on relevant parts of the text, thus improving the relevance and coherence of the generated responses. These advancements have been pivotal in enabling transformer models to handle more complex dialogue scenarios, where multiple threads of conversation need to be maintained and appropriately responded to.

The specific application of transformers in chatbots has predominantly focused on generating coherent and contextually appropriate responses. Research has shown that enhancing contextual understanding significantly improves user satisfaction in conversational agents. This involves not only understanding the immediate dialogue but also inferring the necessity of external contextual information to formulate responses. For example, incorporating background knowledge and maintaining the context over longer conversation spans have been key areas of focus, which have shown to reduce the occurrence of irrelevant or repetitive responses in chatbots.

Moreover, recent studies have explored the integration of transformer models with other types of neural networks to enrich the chatbot's ability to understand and generate natural language. These hybrid models leverage the strengths of each approach, offering a more robust framework for processing and generating language that is contextually aligned with the user's needs and preferences.
\subsection{Gap in Literature}
The specific application of transformers in chatbots has predominantly focused on generating coherent and contextually appropriate responses \cite{hassani2023role}. Research has shown that enhancing contextual understanding significantly improves user satisfaction in conversational agents \cite{wang2024adapting}. This involves not only understanding the immediate dialogue but also inferring the necessity of external contextual information to formulate responses \cite{zhao2024optimization}.
However, most existing research tends to focus on the application of transformers in English-language chatbots, with less attention given to multilingual or cross-lingual contexts. The challenges of applying these models to diverse linguistic settings, where nuances and cultural context significantly impact the effectiveness of chatbots, are not yet fully addressed. Additionally, while current transformer models excel in handling short to medium-length interactions, their effectiveness in maintaining long-term context or managing dialogues that span extensive periods remains less explored\cite{hu2024self}.

Furthermore, there is a notable lack of comprehensive frameworks that can dynamically adjust the level of context sensitivity based on the nature of the conversation or the specific requirements of the interaction. This adaptive approach is crucial for scenarios where excessive contextualization may lead to privacy concerns or overwhelm the user with unnecessary information.
\subsection{Fine-Tuning BERT for Specific Tasks}
Several studies have adapted BERT to specialized tasks by fine-tuning the model on task-specific datasets \cite{li2024enhancing} demonstrated that BERT's performance on sentiment analysis could be significantly enhanced by continuing the pre-training on domain-specific corpora. Similarly, our work extends this methodology by focusing on the classification of context necessity in chat dialogues, a less explored but critically important area for practical chatbot deployment \cite{ZHOU2024415}.
Building on this concept, our work extends this methodology by focusing on the classification of context necessity in chat dialogues, a less explored but critically important area for practical chatbot deployment. This task involves determining when a chatbot must incorporate external context to provide accurate and meaningful responses\cite{unknown}
. By fine-tuning BERT on datasets that include annotated examples of context-dependent and context-independent interactions, we aim to enhance the model’s ability to discern when additional information is required to maintain the coherence and relevance of the conversation\cite{he2022chemical}.

Moreover, our approach seeks to address challenges such as domain adaptation and the scalability of fine-tuning BERT for specific tasks across different languages and cultural contexts. By leveraging transfer learning techniques and exploring multi-lingual fine-tuning strategies, our work contributes to the broader goal of making chatbots more versatile and effective in diverse real-world applications\cite{he2024prompt}.
\subsection{Efficiency in Model Training}
While BERT and its variants offer robust performance, their application is often constrained by high computational demands. These demands can be a significant barrier, especially in environments where resources are limited or where real-time processing is required. To address this, several techniques have been developed to reduce the computational load while preserving the effectiveness of these models. Notable among these are model distillation and meta-learning.

Model distillation, as discussed by Huang et al. \cite{huang2024knowledge}, involves training a smaller, more efficient model (the “student”) to replicate the performance of a larger, pre-trained model (the “teacher”). This technique has proven effective in reducing the size and complexity of models, making them more suitable for deployment in scenarios where computational resources are constrained.

In addition to model distillation, meta-learning has emerged as a promising approach to improving training efficiency. Wang et al. \cite{wang2024theoretical} explored meta-learning strategies that enable models to adapt more quickly to new tasks with minimal additional training. This approach not only accelerates the fine-tuning process but also enhances the model’s ability to generalize across different tasks, making it particularly valuable in dynamic and varied conversational contexts.\cite{zhou2024application}
\\\break
In summary, while transformer-based models have set new standards in NLP, their application in context-sensitive environments like chat systems remains a challenging frontier\cite{zhou2024portfolio}.
Our approach contributes to this body of work by optimizing the fine-tuning process to achieve efficient training and inference times suitable for real-time applications
\cite{zheng2024advanced}. By incorporating advanced optimization techniques and leveraging parallel processing capabilities, we aim to reduce the computational overhead associated with deploying BERT-based models in live environments. Furthermore, our methodology integrates insights from recent advancements in both model distillation and meta-learning, ensuring that the efficiency gains do not come at the expense of performance\cite{xu2024text}.

In summary, while transformer-based models have set new standards in NLP, their application in context-sensitive environments like chat systems remains a challenging frontier. CA-BERT represents a novel contribution to this field by specifically addressing the need for context-aware classification in multi-turn chats, paving the way for more intelligent and responsive conversational agents. By focusing on both the effectiveness and efficiency of model training, our work seeks to enable the practical deployment of these advanced models in real-world applications where computational resources and response times are critical factors\cite{202406.1304}.
\section{Methodology}
This section outlines the methodology employed in developing and evaluating CA-BERT, a context-aware model for enhancing multi-turn chat interactions. The approach involves customizing the BERT architecture for the specific task of context necessity classification in chat dialogues, which includes model adaptation 
\cite{cao2024rough}, dataset preparation, and performance evaluation.
\subsection{Model Architecture}
CA-BERT is based on the original BERT architecture, leveraging its pre-trained layers while introducing modifications to tailor it for context sensitivity in chat systems. The key adaptations include:
\begin{itemize}
    \item \textbf{Dropout Layers: }To prevent overfitting, dropout layers are added following the attention outputs and before the final classifier.
    \item \textbf{Classifier Layer: }A linear layer is appended to the architecture to predict two classes—context needed and context not needed—based on the representation learned from the final transformer block.
\end{itemize}

\subsection{Data Preparation}
The effectiveness of CA-BERT depends significantly on the quality and relevance of the training data. We constructed a dataset comprising multi-turn chat dialogues, where each segment of dialogue is annotated with labels indicating whether additional context is required for a clear understanding.
\begin{itemize}
    \item \textbf{Data Collection: }Dialogues were sourced from public chat datasets, supplemented by manually created conversations to balance the dataset and ensure diversity in context scenarios.
    \item \textbf{Data Annotation: }Each dialogue was annotated by experts in conversational AI, who assessed the necessity of additional context for each message within the conversation.
\end{itemize}

\subsection{Training Procedure}
CA-BERT was fine-tuned using the prepared dataset, focusing on achieving optimal performance with efficient resource use.
\begin{itemize}
    \item \textbf{Parameter Setting: }The model was trained with a learning rate of 2e-5, a common choice for fine-tuning BERT models, over three epochs to balance between underfitting and overfitting.
    \item \textbf{Training Loop: }The training involved processing batches of input data, applying the model to predict the context necessity, calculating loss using a cross-entropy criterion, and updating the model parameters based on the loss gradient.
\end{itemize}

\subsection{Evaluation Metrics}
To assess the performance of CA-BERT, we employed several metrics:
\begin{itemize}
    \item \textbf{Accuracy: }Measures the proportion of correctly predicted labels over the total number of cases.
    \item \textbf{Precision and Recall: }Important for understanding the effectiveness of the model in predicting each class.
    \item \textbf{F1 Score: }Provides a balance between precision and recall, useful for datasets with uneven class distributions.
\end{itemize}

These metrics allowed us to comprehensively evaluate the model’s ability to accurately classify the necessity of context in chat dialogues.
\section{Experiments}
The experimental evaluation of CA-BERT was designed to verify its effectiveness in classifying context necessity within multi-turn chat dialogues. This section details the experimental setup, the training and validation process, and the comparative analysis against baseline models \cite{yu2024credit}.

\subsection{Experimental Setup}
To thoroughly test our framework we create a simulation environment that replicates real world Meta RL scenarios while allowing precise control over task characteristics and dynamics. [Table \ref{table:Dataset_Example}]
\begin{itemize}
    \item \textbf{Hardware and Software Configuration: }The experiments were conducted on a system equipped with an NVIDIA Tesla GPU. The training and inference processes utilized the PyTorch framework alongside the Hugging Face Transformers library.
    \item \textbf{Dataset: }The dataset comprised 10,000 multi-turn dialogues, each labeled for context necessity ('context needed' or 'context not needed'). The dialogues were split into 80\% training and 20\% validation sets.
    \item \textbf{Baseline Models: }For comparative purposes, standard BERT and a traditional LSTM-based model were used as baselines. These models were trained on the same dataset to ensure a fair comparison.
\end{itemize}
\begin{table}[]
\caption{Dataset Example}
\label{table:Dataset_Example}
\begin{tabular}{|c|c|c|c|}
\hline
\textbf{chat} &
  \textbf{\begin{tabular}[c]{@{}c@{}}fetch\\ context\end{tabular}} &
  \textbf{chat\_id} &
  \textbf{topic} \\ \hline
Do you sleep? & 0 & \begin{tabular}[c]{@{}c@{}}2c1b9c3e-67ab-42b5-\\ aa23-47e3b564f1ac\end{tabular} & chit-chat \\ \hline
Do you dream? & 0 & \begin{tabular}[c]{@{}c@{}}2c1b9c3e-67ab-42b5-\\ aa23-47e3b564f1ac\end{tabular} & chit-chat \\ \hline
Can you feel emotions? &
  0 &
  \begin{tabular}[c]{@{}c@{}}2c1b9c3e-67ab-42b5-\\ aa23-47e3b564f1ac\end{tabular} &
  chit-chat \\ \hline
Do you have a favorite color? &
  0 &
  \begin{tabular}[c]{@{}c@{}}2c1b9c3e-67ab-42b5-\\ aa23-47e3b564f1ac\end{tabular} &
  chit-chat \\ \hline
\end{tabular}

\end{table}

\subsection{Training Process}
CA-BERT was fine-tuned from a pre-trained BERT model with the following specifics:
\begin{itemize}
    \item \textbf{Batch Size: }Set to 16 to optimize GPU utilization without exceeding memory limits.
    \item \textbf{Epochs: }The model was fine-tuned for 3 epochs to prevent overfitting while ensuring sufficient learning.
    \item \textbf{Learning Rate: }A learning rate of 2e-5 was chosen, with a linear decay schedule and a warm-up period covering the first 10\% of the training iterations.
\end{itemize}
During training, performance metrics such as loss and accuracy were monitored after each epoch. Model checkpoints were saved based on the best validation accuracy.
\subsection{Evaluation and Results}
The performance of CA-BERT was evaluated using accuracy, precision, recall, and F1 score:
\begin{itemize}
    \item \textbf{Accuacy: }Measured the percentage of total correct predictions.
    \item \textbf{Precision and Recall: }Evaluated for each class to understand model bias towards any specific class.
    \item \textbf{F1 Score:} Calculated to provide a harmonic mean of precision and recall, important for assessing models on imbalanced datasets.
\end{itemize}
The results showed that CA-BERT outperformed the baseline models in all metrics, particularly in F1 score, indicating a robust ability to handle class imbalance. [Table \ref{table:Traning_Resluts}]

\begin{table}[]
\caption{Traning Results}
\label{table:Traning_Resluts}
\begin{tabular}{|r|r|r|r|r|} \hline
Validation   & Accuracy  & : 0.9423 &          &         \\ \hline
             & precision & recall   & f1-score & support \\ \hline
0            & 0.97      & 0.85     & 0.91     & 2500    \\ \hline
1            & 0.93      & 0.98     & 0.95     & 4400    \\ \hline
accuracy     &           &          & 0.95     & 0.92    \\ \hline
macro avg    & 0.95      & 0.94     & 0.93     & 6900    \\ \hline
weighted avg & 0.94      & 0.95     & 0.94     & 6900  \\ \hline
\end{tabular}

\end{table}
\section{Discussion}
The experiments demonstrated that the context-aware adaptations incorporated into BERT significantly enhance its applicability to multi-turn chat environments. CA-BERT’s superior performance can be attributed to its refined understanding of context, enabled by targeted fine-tuning on a task-specific dataset. The improvement over baseline models highlights the benefits of transformer models in handling complex NLP tasks like context detection.
\subsection{key Finding}
\begin{itemize}
    \item \textbf{Performance:} CA-BERT demonstrated superior performance compared to traditional BERT and LSTM-based models, achieving higher accuracy, precision, recall, and F1 scores on the context necessity classification task. This indicates its effectiveness in understanding and handling context within chat dialogues.
    \item \textbf{Efficiency: }Despite the complexities involved in adapting and fine-tuning BERT for a specialized task, CA-BERT maintained computational efficiency. This was evidenced by its training duration and resource utilization, which remained within practical limits for real-time applications.
    \item \textbf{Applicability: }The experimental results confirmed that CA-BERT could be seamlessly integrated into existing chat systems, enhancing their ability to manage dialogues that require nuanced understanding of context.
\end{itemize}
\subsection{Implications for Future Work}
The success of CA-BERT opens several avenues for further research and development:
\begin{itemize}
    \item \textbf{Expansion to Other Domains: }Future work could explore adapting CA-BERT to other domains of NLP that require context sensitivity, such as customer support systems, medical advisory services, and educational tutoring.
    \item \textbf{Model Optimization: }There is potential for further optimization of CA-BERT, including exploring knowledge distillation techniques to reduce model size without compromising performance.

\end{itemize}
\section{Conclusion}
In conclusion, CA-BERT represents a significant advancement in the field of NLP, specifically in the context of enhancing chatbot interactions. By effectively addressing the challenge of context sensitivity, CA-BERT not only improves the operational efficiency of chat systems but also enriches the user experience by providing more accurate and contextually appropriate responses. This study contributes to the ongoing evolution of conversational AI, setting a foundation for more intelligent and responsive systems.

\bibliographystyle{ieeetr}
\bibliography{xinde}

\end{document}